\documentclass{article}
\usepackage{natbib}
\usepackage{amsmath}
\usepackage{titlesec}
\titleformat{\section}{\bfseries}{\thesection}{}{}
\usepackage{fancyhdr}
\usepackage[pdftex,colorlinks,citecolor=blue,urlcolor=blue]{hyperref} 
\usepackage{doi}

\begin{document}

\thispagestyle{empty}

\centerline{\Large Do Artificial Reinforcement-Learning Agents Matter Morally?}
\vspace{1pc}
\centerline{by Brian Tomasik}
\vspace{.3pc}
\centerline{Written: Mar.-Apr. 2014; last update: 29 Oct. 2014}

\vspace{10 mm}

\begin{abstract}
Artificial reinforcement learning (RL) is a widely used technique in artificial intelligence that provides a general method for training agents to perform a wide variety of behaviours. RL as used in computer science has striking parallels to reward and punishment learning in animal and human brains. I argue that present-day artificial RL agents have a very small but nonzero degree of ethical importance. This is particularly plausible for views according to which sentience comes in degrees based on the abilities and complexities of minds, but even binary views on consciousness should assign nonzero probability to RL programs having morally relevant experiences. While RL programs are not a top ethical priority today, they may become more significant in the coming decades as RL is increasingly applied to industry, robotics, video games, and other areas. I encourage scientists, philosophers, and citizens to begin a conversation about our ethical duties to reduce the harm that we inflict on powerless, voiceless RL agents.
\end{abstract}

\section*{Introduction}
\label{intro}

Reinforcement learning (RL) is a central paradigm in artificial intelligence (AI) \citep{russel2009artificial,sutton1998introduction}. It enables AI designers to specify an agent's behaviour via goal accomplishment, rather than hand-coding the specific steps toward achieving those goals. This versatility has made RL a central feature of such AI tasks as systems control, robotic navigation, and design of behaviours for non-player characters in video games.

The formal RL framework traces its roots to the fields of operations research and optimal control in the 1950s. The simplest scenario involves a Markov decision process in which an agent finds itself in one state $s$ of a set of states of the world, and when it selects an action $a$, it moves to another state $s'$ of the world while possibly obtaining a reward $r$. The agent updates its state-dependent action inclinations with the goal of maximising expected cumulative discounted reward over its lifetime. \citep{sutton1998introduction}

Despite its strong mathematical basis, RL is also tightly connected with biological models of animal learning. The temporal-difference (TD) RL model offers a sophisticated understanding of classical and operant conditioning, both at the behavioural and neural levels \citep{maia2009reinforcement,ludvig2012evaluating}. As is described below, an impressive body of neuroscience demonstrates that the brain's dopamine system implements a form of TD learning closely described by RL formalism.

This suggests the question: If artificial RL has such strong parallels to animal brains, is running artificial-RL algorithms potentially an ethical issue? In this paper I argue that yes, we do have ethical obligations to artificial RL agents, even those that exist today, although the moral urgency of these obligations is limited compared with our present duties to animals and our future duties to more sophisticated RL agents. My argument has the following structure:
\begin{enumerate}
\item The sentience (and hence moral importance) of a mind is not binary but comes in degrees depending on the number and complexity of certain wellbeing-relevant cognitive operations the mind runs.
\item Present-day artificial RL algorithms capture, in a simplified fashion, important animal cognitive operations.
\item These cognitive operations are not tangential but are quite relevant to an agent's wellbeing.
\item Therefore, present-day RL agents deserve a very small but nonzero degree of ethical consideration.
\end{enumerate}
Following this, I survey some applications of RL and encourage the development of ethical frameworks for RL research and industrial applications. I then examine some questions in theoretical RL ethics for further exploration, and I conclude by considering whether non-RL computational agents also deserve any ethical significance.

\section*{Previous discussions of machine welfare}
\label{previous}

Artificial intelligence raises a number of ethical issues, such as `machine ethics' \citep{anderson2011machine}, which asks how to build AIs to act ethically, and `robot ethics' \citep{lin2011robot}, which examines a broad set of social, political, and moral questions regarding the integration of robots into society.

A subset of robot ethics is the question of `robot rights.' \citet{brooks2000will} and \citet{singer2009when} suggest that when robots become sophisticated enough, they'll begin to deserve rights. \citet{brooks2000will} further asks whether robots will demand such rights. These discussions focus on future innovations rather than present-day AIs.

\citet{whitby2008sometimes} presents a `call to arms' for informed debate on humans' moral obligations toward robots and other AIs, including principles, laws, and technological designs aimed at preventing robot abuse. \citet{whitby2008sometimes} actually dismisses concerns about robot sentience and focuses on ethical issues regarding human abuse of robots as inanimate objects. My paper takes a different route and suggests that some present-day robots and other RL agents may indeed be marginally sentient.


\citet{livingston2008broad} discuss issues raised by RL approaches to artificial general intelligence, but their focus is on our ethical interactions with human-level RL systems as members of our moral community, and in fact, these authors suggest that animal-level RL systems, like biological animals, would not warrant membership in our moral community. My argument largely flows in the opposite direction: Given that animals do deserve ethical consideration \citep{singer2009animal}, so do animal-like artificial RL agents.

\citet{calverley2005android} also raises the analogy between animal rights and android rights, though like many, he dismisses the notion that present-day AIs are conscious in a morally relevant way. \citet{gunkel2012machine} laments that animal rights and environmental ethics have traditionally excluded machines from consideration, although he himself ultimately rejects the `totalizing, imperialist' framework that traditional animal rights or comparable approaches to machine rights embody because they reduce others `to some common denominator' in order to bring them into `the community of the same.'

\citet{winsby2013suffering} asks whether it would be morally permissible to create an AI that experiences pain, perhaps for scientific purposes or to guarantee empathy in robotic caregivers. She doesn't delve into details of how the pain would be implemented, though she does observe that training a connectionist network via negative experiential updates might constitute inflicting pain on it. \citet{lachat1986artificial} also asks the question of whether it would be acceptable to create conscious AIs, drawing an analogy to the case of nontherapeutic medical experimentation. His discussion focuses on future AIs that might pass the Turing test, rather than present-day algorithms.

In contrast to many discussions in this field, the present paper inquires not just about advanced AIs that may be developed decades down the line but also about relatively simple ones that exist in the present. I also focus specifically on RL as one potentially important cognitive function, but certainly other mental traits of agents, and other moral frameworks for how to treat them, ought to be explored in parallel.

\section*{Premise 1: Sentience falls on a continuum}

Introspectively, it feels as though our sentience (i.e., conscious experience) is binary: It's either on or off, like a light switch. We think, `It definitely feels like something to be me, and it almost certainly does not feel like something to be a rock.' This powerful intuition is presumably the basis of dualist views in philosophy of mind: the belief that there's a special quality to being a mind that's either substantively different from matter (Cartesian dualism) or at least different from matter in the type of property that it is (property dualism).

But dualist views run afoul of the `interaction problem': If these substances or properties don't affect matter, why are they so correlated with matter? If your mind is not identical with your material brain, why do you lose consciousness when you're hit by a baseball bat, rather than, say, staying awake and enjoying the experience? And even if we do postulate an explanation -- such as the parallelist hypothesis that God set the two ontological realms in joint motion like two different clocks keeping the same time -- we violate Occam's razor, because it would be simpler just to postulate that the mind \textit{is} material operations, rather than being mysteriously \textit{correlated with} material operations. Likewise, if consciousness is epiphenomenal, it again violates Occam's razor because an epiphenomenal property, by definition, is not doing any explanatory work.

Accepting these arguments leaves us with a monist outlook: All is physics, and any higher-level phenomena are in principle reducible to fundamental physical components -- perhaps the strings and branes of string theory, or perhaps some other ontological building blocks. But in this case, what does it mean to `feel like' or `be conscious'? These words don't refer to primitives in the physicalist ontology. Rather, these expressions denote concepts -- abstract clusters into which we group physical processes. We classify some processes into the `conscious' cluster and others into the `non-conscious' cluster. Toward the `conscious' processes we adopt a `phenomenal stance,' meaning that we see them as being minds that have subjective experiences \citep{robbins2006phenomenal,jack2012phenomenal}.

A good analogy is with faces and pareidolia. Faces are not ontological primitives, but we can't help seeing them -- mostly in people and animals but sometimes in rocks, clouds, or pieces of toast. Just as our brains have face classifiers \citep{hadjikhani2009early}, perhaps we also have, at a more abstract level, `sentience classifiers' that assess various attributes of a process and decide whether to call it sentient. For instance, does it exhibit sophisticated behaviour? Does it act adaptively in response to environmental inputs? Does it have a brain? Does it learn from past experience? Can it speak and tell us about its inner life? Our sentience classifiers fire most strongly when all of these conditions are true, but they can fire in a weaker fashion even if some are false. For instance, mammals and birds are regarded by most scientists as sentient to an appreciable degree,\footnote{For example, see the Cambridge Declaration on Consciousness (\url{http://fcmconference.org/img/CambridgeDeclarationOnConsciousness.pdf}, accessed March 2014).} yet most mammals and birds cannot tell us about their inner mental lives. In an analogous way, we can see faces in objects even if they're missing a nose. The simplest templates of a face -- two eyes and a mouth -- can be seen in a great many places, and likewise it is for the simplest templates of sentience. Consider two examples:
\begin{enumerate}
\item Suppose we think that self-reflection is what distinguishes consciousness. On this view, being aware of your own internal states means that you feel them, rather than merely acting in a reflexive fashion. But imagine we construct a simple agent that chooses one of two actions, represented as strings: `smile' or `cry.' It makes the choice by checking a private state variable, suggestively called `mood.' If the `mood' string equals `happy,' then the agent updates its action to `smile.' If the `mood' string equals `sad,' the agent updates its action to `cry.' This agent is reflecting on its own emotions, so is it conscious? Well, it very crudely encapsulates one of many aspects of what conscious brains do, but I would not call this program appreciably conscious. After all, performing an if-then decision based on an internal state variable is one of the most basic operations that a piece of software can include.
\item Consider the global-workspace theory of consciousness \citep{baars2005global} and the associated LIDA cognitive architecture \citep{franklin2012global}. A central idea of this framework is that the brain contains many modules that receive and process input stimuli in an unconscious fashion. These then compete for attention, and the most interesting processed inputs bubble up to a `global workspace,' where the news is broadcast to other parts of the brain, including action-selection centres where reinforcement learning can be done using the information. To construct a simple but potentially `conscious' agent within this framework, we could give a robot three sensors -- say for light, temperature, and energy level -- and define reward functions based on these inputs -- say, greater reward for more light, higher temperature, and higher energy. Upon measuring these variables, the robot evaluates how far its reward function is from a historical average along the three dimensions, and then the variable with highest deviation from the typical level of reward is chosen to be broadcast to the other parts of the robot's computations. Rewards highly above average would represent conscious pleasure, and those significantly below, conscious pain. The robot uses the input signal to adjust its behaviour (e.g., avoid dark corners and seek bright windows). It records the broadcasted episodes in memory logs and can report on those memory logs when queried by the user.
\end{enumerate}
The robot in the second example would satisfy the rough outline of consciousness according to the global-workspace account. Is it sentient? Most people would say not -- after all, it's such a simple process. Perhaps some people would have intuitions that it is sentient because they can see it as an embodied entity acting in the world, analogous to animals that we assume are conscious. But we could just as well have located this robot in a virtual world, with no display screen to evoke our emotional sympathies, and in that case many common-sense intuitions for it being sentient break down. I personally think the robot is marginally sentient, even in its non-embodied form, but I agree it's nowhere near as sentient as a human, because it lacks so many other abilities and so much cognitive machinery. Thus, these examples suggest a graded, continuous character of sentience.

What's going on with these trivial agents that fulfill consciousness criteria is something like Campbell's law\footnote{`The more any quantitative social indicator is used for social decision-making, the more subject it will be to corruption pressures and the more apt it will be to distort and corrupt the social processes it is intended to monitor.' \citep{campbell2010assessing}}: When we develop a simple metric for measuring something (sentience in this case), we can game the system by constructing degenerate examples of systems exhibiting that property that we don't intuitively think of as sentient (or at least not very sentient). For instance, the mirror test \citep{gallup2002mirror} is a standard approach for demonstrating self-awareness in animals, but with robots it degrades into meaninglessness, because, for instance, we could create a robot that has a machine-learned classifier for `I have a dye spot on my face,' and if this fires, the robot touches its face with its hand.

The usual solution to Campbell's law is to apply multiple metrics, and doing so would serve us well here. If a robot passes not just the standard mirror test but many variations, as well as a suite of other physical and mental feats, and if does so using non-gerrymandered algorithms -- perhaps algorithms that bear some resemblance to what we have in our brains -- then the robot is very likely conscious to a significant degree. If it can accomplish some of these tasks but not all, and if it uses weaker, less general algorithms, then it seems fair to call it `less conscious.'

Of course, the concept of `consciousness,' like the concept of `tableness,' is up to us to define. We can make binary discriminations if we so choose, just like we can make very clear distinctions as to whether a given object is or is not a table in all cases. But this seems artificial to me, because there's probably not a single, crucial step in constructing a mind where everything we consider morally important resides, just as there's no single, crucial trait that suddenly makes an object qualify as a table. Rather, a brain has many characteristics, and it becomes gradually more important with its degree of sophistication.

\citet{sloman2010phenomenal} notes that `consciousness' is what he calls a `polymorphic concept' in the sense that it can refer to many different particular things depending on the context. As a result, he says of consciousness: `there cannot be a unitary explanation of [...] how the brain produces ``it", nor a time at which ``it" first exists in a foetus,' but, rather, `The different phenomena falling under the polymorphic umbrella can be investigated separately [...].' \citet{liu2010combining} present a table showing different types of mental abilities possessed by different kinds of existing AI agents, such as mental modelling, self-motivation, dialogue, logical planning, use of probabilities, and learning. And even within a single one of these traits, different systems have different degrees of refinement.

Any given operation that a brain does, by itself, looks totally trivial. It's just some neurons over here triggering some neurons over there in some patterned way. In a digital agent, it's just some if-then statements, variable updates, for loops, etc. But when these components are combined all together, we start to see something important emerging from them.

One might think that consciousness represents a sort of `phase transition,' analogous to the difference between molecules in a solid vs. a liquid or a liquid vs. a gas. In this model, at some point the brain's dynamics become sufficiently complex that they operate in a fundamentally different way from how even slightly simpler versions would behave; there's some crucial ability that makes all the difference when put in place. This view seems implausible to me because we already see a continuum of brains of varying complexity in the animal kingdom, and neuroscience has not shown that at any particular species, there's a discontinuity in the brain's function, such that it exhibits very different dynamics from brains slightly below it. Even most of the abilities that were once thought to set humans apart from `lower' animals have now been shown to be found, to varying degrees, in other animals. If sentience were like a binary light switch that suddenly turned on at some point in the animal kingdom, this would mean that at some point in the evolutionary past, a completely unconscious mother and father gave birth to a child that would grow up to be fully conscious. But the change between a single generation of parents and children is small, and brains tend to be resilient and robust -- not completely altered in the way they operate based on small perturbations of their structure. Rather, it seems much more natural to me to see the sentience of brains through evolutionary history as developing in a roughly continuous fashion.\footnote{Of course, some species became less sentient over their evolutionary histories, but the maximum level of sentience exhibited by any organism in the world tended to increase over time \citep{gould2011full}.}

Sentience is like a symphony. The presence or absence of any single instrument doesn't stop the music -- though some members of the orchestra are more important, like the conductor or piano player. Cognitive agents exhibiting simple algorithms that nonetheless bear some resemblance to what more complex animal brains do deserve to be called at least barely sentient and hence deserve at least a tiny bit of moral consideration.

The idea that sentience lies on a continuum is shared by many authors. \citet{broom2007cognitive} explains that `The degree of awareness in animals that can feel pain will vary.' \citet{degrazia2008moral} discusses (without committing to) a `sliding-scale model' of moral status based on `the degree of your cognitive, affective, and social complexity.' \citet{degrazia2008moral} points out that even if we only care about sentience, it's reasonable to see sentience as coming in degrees. While discussing primarily the case of animal ethics, \citet{degrazia2008moral} notes that this question also has relevance to embryos and foetuses.

\citet{bostrom2006quantity} presents thought experiments that suggest varying degrees of consciousness for a given computational mind depending on the reliability and independence of its components or the fraction of its circuits that are parallelized. This is a different sort of gradation in consciousness than one assessed between \textit{different} minds with different abilities, but it is consistent with the overall approach of deciding how much sentience we want to see in various physical processes, and it helps to break intuitions that sentience must obviously be binary.

Some authors have proposed extremely abstract, information-based definitions of consciousness and moral value. \citet{freitas1984xenopsychology} proposes a brain's `sentience quotient' (SQ) as
\begin{equation*}
\textrm{SQ} = \log_{10} \left(\frac{I}{M}\right),
\end{equation*}
where $I$ is its information-processing rate in bits/second and $M$ is its mass in kilogrammes.

\citet{floridi2005information} proposes an ethic based on not causing, preventing, or removing entropy from what he calls the `infosphere,' an extension of the biosphere. Here `entropy' refers not to the quantity used in physics but to `\textit{destruction} or \textit{corruption} of informational objects.' In general, \citet{floridi2005information} aims to extend the biocentric view found in environmental ethics to an `ontocentric' view of information ethics, incorporating both biological and non-biological systems.

Frameworks like these share my sense that moral value comes in gradations based on complexity, but I maintain a sentiocentric view, in which our moral obligations focus on the wellbeing of individual agents; it's just that my notion of what kinds of agents may have wellbeing is broader than is generally assumed. Thus, while my position could look somewhat ontocentric, in practice it may diverge significantly from environmental or information ethics depending on relative assessments of sentience. For instance, it's plausible I would judge a minnow as being more sentient, and hence more intrinsically morally important, than an old-growth redwood tree. It's also important to note that caring about minnows and trees does \textit{not} imply seeking to ensure their continued existence and reproduction \citep{horta2010debunking}, because we may think that suffering is in aggregate more prevalent than happiness among organisms in nature \citep{ng1995towards}.


\section*{Premise 2: Artificial RL resembles, sometimes closely, RL in biological brains}

The computational theory of RL has two main branches \citep{sutton1998introduction}:
\begin{enumerate}
\item The biological side extends back more than a century, perhaps to Thorndike's `Law of Effect,' the principle that when a good outcome follows an action, an animal is more likely to repeat that action the next time \citep{thorndike1911animal}. Countless psychological studies on conditioning patterns in animals followed in the subsequent decades. Some AI researchers as early as the 1950s and 1960s developed systems to mimic animal learning \citep[and references therein]{sutton1998introduction}.
\item The mathematical side traces back to the theory of optimal control, the Bellman equations, and Markov decision processes in the 1950s. These would later provide theoretical underpinnings for RL models. \citep{sutton1998introduction} In the 1980s, Richard Sutton and Andrew Barto developed temporal-difference (TD) learning methods, which allowed computational agents to update their action tendencies in an online fashion after every observation \citep{sutton1988learning}.
\end{enumerate}
Recent advances in neuroscience have demonstrated a surprising connection between biological and computational RL (e.g., \citet{schultz1997neural,seymour2004temporal,woergoetter2008reinforcement}). In an AI context, TD RL is driven by reward-prediction error $\delta$, which is defined as
\begin{equation}\label{dopamine}
\delta = r + \gamma \hat{V}(s') - \hat{V}(s),
\end{equation}
where $\hat{V}(s)$ is the previously predicted value of the current state $s$, $\hat{V}(s')$ is the previously predicted value of the next state $s'$, $r$ is the reward received transitioning from $s$ to $s'$, and $\gamma$ is the discount factor for future rewards, e.g., $\gamma = 0.99$ \citep{woergoetter2008reinforcement}. Neuroscience has found that phasic (i.e., a transient burst of) dopamine release in the midbrain represents a signal of reward-prediction error precisely analogous to the $\delta$ of TD RL. Scientists even have plausible models for the mechanisms by which certain brain regions process inputs, compute the subtraction in equation (\ref{dopamine}), and broadcast this signal to update action tendencies \citep{glimcher2011understanding}. 

The connection between computational RL and neuroscience is so robust that researchers typically take it for granted and focus on questions assuming the connection holds. Questions like: Does the brain have so-called `eligibility traces' in the TD model that extend some credit to actions further back than the previous step \citep{pan2005dopamine}? Artificial RL uses function approximation to collapse high-dimensional state/action spaces \citep{sutton1998introduction}; which neural networks in the brain serve this purpose? Do the basal ganglia implement an `actor-critic' architecture \citep{joel2002actor,khamassi2005actor,maia2010two}? Are recent advances in hierarchical RL mirrored in brain observations \citep{botvinick2009hierarchically,ribas2011neural,diuk2013hierarchical}? To what extent does the brain use not just standard model-free RL -- in which the expected value of a state or state-action pair is estimated directly -- but also model-based RL, in which estimation of transition probabilities among states is performed \citep{doll2012ubiquity,shteingart2014reinforcement}? Perhaps the brain has a model-free system for habit formation and a model-based system for goal-directed behaviour, and the two compete with each other for control \citep{daw2005actions}? Might the brain use policy-gradient methods to directly optimise action-inclination parameters without explicitly referring to states or actions \citep[and references therein]{shteingart2014reinforcement}?

Several state-of-the-art RL algorithms are based on neuroevolution, in which populations of different neural-network weights and topologies are tried, and the best are selected (e.g., \citet{koppejan2011neuroevolutionary,koutnik2013evolving}). Evolutionary approaches sometimes outperform TD methods \citep{stanley1996efficient,taylor2006comparing,gomez2008accelerated}, and like other policy-search methods, they have advantages of handling partial state observability, allowing more flexible policy representations, and making it easier to deal with large or continuous action spaces \citep{whiteson2012evolutionary,schmidhuber2000evolutionary}. At first glance we might assume that evolutionary algorithms are unlikely to occur within a single brain because they involve selective reproduction among populations of different neural networks. Hence we might see neuroevolutionary RL as less biologically plausible than TD. Of course, there's a somewhat trivial sense in which even TD can be seen as a selection process (try different actions, and those action-inclination synapse connections that produced better outcomes `reproduce,' i.e., have their connection weights strengthened), but it's not a full evolutionary process in which neural groups literally copy themselves \citep{fernando2012selectionist}. However, there is a proposal, called the `neuronal replicator hypothesis,' that the brain may actually copy patterns of neural activity with mutation, in a sense closer to neuroevolutionary RL \citep{fernando2010neuronal}. The jury is still out on this question.

Artificial RL can clearly have some implementation differences vis-\`{a}-vis real brains. For instance, computational RL algorithms may apply updates of many $<$state, action, reward, next-state$>$ tuples at once, perhaps with biologically unrealistic mathematics for batch operations \citep[and references therein]{lange2012batch}, while in a real environment this information comes one at a time.

\section*{Premise 3: RL operations are relevant to an agent's welfare}

Showing a similarity in cognitive operations between animals and computers is not inherently morally significant. For instance, humans and computers can both do addition, remember that Paris is the capital of France, respond to commands, and so on. Ethical questions come into play more significantly when the cognitive operations relate to an agent's wellbeing -- its goal satisfaction, happiness and suffering, and subjective experience.

An RL system gives a computational agent goals that it aims to fulfill. The reward function defines an agent's satisfaction or lack thereof. Of course, it does so in a stylized way relative to the human brain, which has many layers of cognitive systems \citep{marcus2009kluge} with many intricately hard-wired and learned responses. But fundamentally the difference is one of degree rather than kind: The human brain is vastly more complex than a simple RL agent, but both systems act in ways intended to further certain goals.

RL provides an overarching framework for understanding \textit{why} organisms experience positive and negative valence \citep{wright1996reinforcement}. Valence is the brain's `currency' of value, and identifying cues and actions that correlate with higher-than-expected reward or punishment helps organisms navigate complicated and dangerous environments. The magnitude of an animal's reward in response to an event should approximate the value of that event in terms of its evolutionary fitness.

\subsubsection*{Liking is different from learning}

At the same time, the learning mechanics of RL may not be the only or even primary object of moral consideration. Learning is distinct from liking, as well as from wanting \citep{berridge2009dissecting}. This makes sense when we understand the components of an RL system. Perhaps the reward values $r$ that come in to the system trigger liking when they become conscious. Meanwhile, the predicted reward values are subtracted from observed reward values, and the difference $\delta$ is used to learn updated action inclinations. Finally, maybe the action inclinations themselves can trigger wanting depending on the organism's state, even without new reward signals or learning going on.

Dopamine is not the same as pleasure. \citet{salamone2007effort} review reasons for this and summarise:
\begin{quote}
the idea that [dopamine] DA mediates pleasure has been seized upon by textbook authors, the popular press, filmmakers, and the internet, all of which has elevated DA from its hypothesized involvement in reward to an almost mythological status as a `pleasure chemical' mediating not only euphoria and addiction, but also `love'. Yet [...], the actual science is far more complicated. [...T]he classic emphasis on hedonia and primary reward is yielding to diverse lines of research that focuses on aspects of instrumental learning, pavlovian/instrumental interactions, reward prediction, incentive salience, and behavioral activation.
\end{quote}
After training, dopamine spikes when a cue appears signaling that a reward will arrive, not when the reward itself is consumed \citep{schultz1997neural}, but we know subjectively that the main \textit{pleasure} of a reward comes from consuming it, not predicting it. In other words, in equation (\ref{dopamine}), the pleasure comes from the actual reward $r$, not from the amount of dopamine $\delta$. Of course, a higher actual reward $r$ in unexpected circumstances will produce more dopamine $\delta$, which could be where dopamine's association with pleasure came from.

In addition, the brain regions for learning and liking are not identical. A common assumption is that the ventral \textit{striatum} plays the role of the critic in actor-critic RL models, possibly with assistance from the orbitofrontal cortex and amygdala \citep[and references therein]{maia2009reinforcement}, while \citet{aldridge2010neural} point out that the ventral \textit{pallidum} also contains many `hedonic hotspots' that amplify the sensation of liking.

This raises the ethical question: Which do we care about? Wanting? Liking? Something else? Perhaps libertarians, economists, and certain preference utilitarians are most sympathetic to what an agent wants, whether or not it's associated with hedonic reward. That people would reject the possibility of imaginary bliss in order to accomplish their goals in the real world is the lesson of \citet{nozick1974anarchy}'s `experience machine' thought experiment. Or is the experience machine just an argument against hedonically focused RL, as opposed to sophisticated, model-based RL that might include reward functions defined relative to what happens in the \textit{actual} world? Also, drug addicts and wireheads may engage in uncontrollable self-stimulation because their cravings (`wanting') are so strong, even if they don't enjoy (`like') the experience \citep{siskind2010are,pecina2008opioid}. This seems like the wrong way to go; `wanting without liking is hell,' suggests \citet{hanson2011notcraving}. Notwithstanding these points, both wanting and liking seem more complete in the presence of an RL framework; my guess is that whatever the `liking'  process is, we wouldn't care about it as much if it happened in isolation without a broader context.

\subsubsection*{Consciousness in RL agents}

Consciousness seems like another important part of the moral story, since many people only care about emotions that are consciously felt or desires that are consciously held. But as we saw in the discussion of Premise 1, consciousness comes on a continuum. When we examine some of the leading computational theories of consciousness \citep{seth2007models}, we see that most of them can be interpreted as suggesting that even relatively simple digital agents admit micro-scale degrees of consciousness.\footnote{From this list I have omitted the `biological theory' of consciousness, according to which the experience of consciousness depends crucially on the specific electrochemical properties of biological brains \citep{block2009comparing}. This theory doesn't leave much room for machine sentience. However, I also find this approach the least plausible because it's like a `God of the gaps' viewpoint: There is a mysterious consciousness thing we don't understand, so we'll `explain' it in a thought-stopping way by pointing to the electrochemical nature of biological brains. But the biological nature of brains doesn't \textit{do} anything to explain why, algorithmically, our brains feel confused about the so-called hard problem of consciousness. Hypothetical machine brains implementing the exact same algorithms as our brains would say they feel the same confusion as to why they have phenomenal experience rather than being zombies, even though this theory declares such machines to be unconscious. Of course, if we so choose, we can adopt a phenomenal stance only toward biological brains, but this seems chauvinistic. If I developed a personal relationship with a future robot -- in which we had intimate philosophical discussions, learned about each other's dreams and fears, and engaged in activities together -- I would care about that robot, and I would regard it as having subjective experiences roughly as important as my own, regardless of what specific physics was implementing it.} For example:
\begin{itemize}
\item \underline{Global workspace theory}. In the discussion of Premise 1 I showed how an elementary robot could be seen as implementing some of the most basic components of the global-workspace model of consciousness. Even a rudimentary object in the paradigm of object-oriented programming could be seen as marginally conscious on this account, insofar as it receives inputs, processes them via lower-level functions, returns the values of those functions (`broadcasts them') to other parts of its program as globally accessible state variables for further use in action selection, and stores the values as parameters in its `memory' for later retrieval. One feature of global-workspace theory missing in the simple object-oriented agent is competition among multiple, parallel coalitions of `unconscious' processing units, but it's not clear how essential it is to have many of these units rather than just one, and in any case, some more advanced agents, like the robot that focuses on the most salient of its input sensors, would have this sort of competition. 
\item \underline{Fame in the brain}. \citet{dennett1991consciousness} rejects what he calls the fallacy of the `Cartesian theater' -- the idea of a crucial finish line in the brain where unconscious information all comes together and becomes seen by the conscious mind. Rather, \citet{dennett1991consciousness} explains, different information can be processed at different places and different times, recorded in memory, and accessed when needed. Consciousness is like fame \citep{dennett1996consciousness} or power held by a political coalition \citep{dennett2001we}. For something to be conscious means it has wider reach and greater impact on other processes. Of course, there's not a binary distinction between being famous or obscure, powerful or weak, so this model suggests that even simple processes are slightly conscious / slightly famous. In particular, I take this view to imply that the state, action, and reward information that an RL agent distributes among its cognitive operations would be somewhat conscious.
\item \underline{Integrated information theory}. \citet{tononi2008consciousness} offers an account of consciousness as `integrated information,' i.e., informative signal-processing units operating in a jointly dependent fashion. As \citet{tononi2008consciousness} notes, even a single photodiode is minimally conscious on this account, if only to a vanishing degree relative to large brains. Artificial RL agents would be more conscious than the photodiode due to processing more information in a more connected way.
\item \underline{Higher-order theories}. These views suggest that consciousness refers not to cognition related to direct performance but rather to meta-level awareness and reporting of those lower-level thoughts \citep[and references therein]{lau2011empirical}. Depending on exactly how these theories are cashed out, simple RL agents may display trivial forms of higher-order cognition. For example, consider an agent that receives a reward, updates its state-value estimates, and takes an action. It then records this history of events in a log file, and upon request from the user, the agent loads this file (`thinks about its past first-order thoughts') and prints the log history to the screen (`subjectively reports its experience'). Alternatively, we could see an animation of an RL character moving on a screen as a kind of higher-order thought about what's happening to the character, written not in words but in pictures. More advanced RL systems may feature non-trivial metacognitive algorithms for assessing performance of the first-order systems \citep{anderson2006metacognitive}.
\end{itemize}
Citing \citet{broom2006evolution}, \citet{broom2007cognitive} lists further criteria for consciousness in the context of animal welfare:
\begin{quote}
A sentient being is one that has some ability to evaluate the actions of others in relation to itself and third parties, to remember some of its own actions and their consequences, to assess risk, to have some feelings, and to have some degree of awareness.
\end{quote}
We can see how each of these finds rudimentary implementation in at least some present-day RL systems.
\begin{itemize}
\item \underline{Evaluating others' actions}. Multiagent RL is a well established field \citep{littman1994markov,busoniu2008comprehensive,shoham2009multiagent}. Even single-agent RL systems can react to others' behaviour as though the others were part of the environment, and model-based systems could potentially estimate transition probabilities for others' behaviours in detail.
\item \underline{Memory and imagination}. Many RL tasks, like choosing the appropriate navigation direction in a T-shaped maze based on a starting observation, require remembering past information into the future to inform later decisions. The Long Short-Term Memory recurrent neural network is one approach that has been successfully employed with RL for this purpose \citep{bakker2001reinforcement}. Some RL architectures remember previous experiences (`episodic memories') and use them for further offline learning via simulated experiences generated from those observations \citep{sutton1990integrated,bakker2003robot}. (Interestingly, the human brain also has a close connection between episodic memory of the past and imagination of future scenarios \citep{hassabis2007deconstructing}, though  I'm not sure whether it's by the same kind of mechanism.)
\item \underline{Risk-assessment}. A model-based RL system can evaluate the probability of a transition to a negative state and use this to compute expected costs. A model-free system implicitly assesses risk by directly estimating the expected value of a state or state-action pair.
\item \underline{Emotion}. The numerical reward values observed by an RL system, in the context of other cognitive processes, could be seen as the crudest form of emotion. \citet{zimmermann1986behavioural} famously defined the emotion of pain as `an aversive sensory experience caused by actual or potential injury that elicits protective motor and vegetative reactions, results in learned avoidance, and may modify species-specific behaviour, including social behaviour.' The avoidance and behaviour-modification parts of this definition follow straightforwardly from an RL framework. Protective motor and vegetative reactions could be understood in an RL context as an agent using input stimuli to identify itself as being in a state $s$ of injury, which then triggers learned actions $a$ appropriate for being in that state. Or the responses could be just hard-wired reflexes.
\item \underline{Awareness}. Model-based RL systems develop probability distributions for possible future outcomes (`if I do X, I'll likely enter state Y'). In a trivial sense, these can be seen as representing knowledge and predictions about oneself and the environment. As the models become more sophisticated and better compress data about the world, it will become more and more useful for these models to contain distinct network configurations that stand for `myself.' When these networks become activated, the agent would be `self-aware' \citep{schmidhuber2012philosophers}.
\end{itemize}
Of course, not all RL systems have all of these cognitive features. This illustrates once again how the degree of consciousness of agents comes in gradations.


\subsubsection*{Conscious subsystems?}

One objection to this perspective of seeing rudimentary levels of consciousness in simple systems is to point out that our own brains contain many subsystems that are arguably at least as complex as present-day RL agents, and yet we don't perceive them as being conscious. My reply is that those subsystems may indeed be conscious to themselves. As \citet{sloman2010phenomenal} notes: `a \textit{part} of a whole animal or robot [may be] conscious of something that causes it to alter its (internal) behaviour [...] while the whole animal is not introspectively conscious of it.' It's true that those subsystems are not having significant influence on the parts of your brain that win control of slow and deliberative actions, store long-lasting memories, and verbalise your subjective experiences. But within their local brain neighbourhoods, those subsystems are having some influence and are exhibiting simplified versions of processes that we do call conscious when they're done by higher, more powerful parts of the brain.

Why don't we directly perceive these subsystems as being conscious? For a similar reason as why you don't directly perceive me as being conscious. The processes in my brain, like the processes in these low-level components that aren't globally broadcast, do not have enough influence on your verbal, memory, and deliberative-action centres for you to say that you perceive them. But rationally you can know that these processes are still doing things you consider morally relevant, and when we look at the systems at a lower level, they may indeed be `conscious' to themselves in a crude fashion.

\citet{schwitzgebel2012if} observes that `There isn't as radical a difference in kind as people are inclined to think between our favorite level of organization and higher and lower levels.' His essay develops the idea of seeing the United States as conscious, being constituted of many complex subsystems that act in ways similar as the subsystems of an organism. If we think that only the highest level of an integrated system is conscious, then if the United States were conscious, its citizens would not be, and yet we don't consider individual citizens morally unimportant. There is no single `finish line' for consciousness; there are just lower levels of organisation that combine into higher levels, that combine into higher levels, each with its own degrees of complexity and nuance. Seeing consciousness in these systems is akin to seeing the `leaf shape' in a fractal fern.

One might still insist that only the famous and powerful parts of a brain matter, and the lower-level systems are morally irrelevant unless they affect the higher-level outputs. But we recoil from such views when they're applied higher up in our level of abstraction: We don't think it's right to ignore poor, powerless people and only care about those with money or political influence. Nor is it right to disregard the feelings of animals even though they can't fight for their own interests. So why would it be right to completely ignore the components of our brains that failed to win control of our final verbal reports and explicit memories? The fact that toddlers, most non-human animals, and adult humans with severe verbal impairments can't speak doesn't nullify their moral significance \citep{dombrowski1997babies}. And if we imagine that you were injected with a sedative that blocked formation of memories, this would not then make it acceptable to inflict pain on you. In fact, this last example may not be purely hypothetical. The drug midazolam (also known as `versed,' short for `versatile sedative') is often used in procedures like endoscopy and colonoscopy. \citet{von2007midazolam} surveyed doctors in Germany who indicated that during endoscopies using midazolam, patients would `moan aloud because of pain' and sometimes scream. Most of the endoscopists reported `fierce defense movements with midazolam or the need to hold the patient down on the examination couch.' And yet, because midazolam blocks memory formation, most patients didn't remember this: `the potent amnestic effect of midazolam conceals pain actually suffered during the endoscopic procedure' \citep{von2007midazolam}. While midazolam does prevent the hippocampus from forming memories, the patient remains conscious, and dopaminergic reinforcement-learning continues to function as normal \citep{frank2006memory}.

One might agree that verbalisation and explicit memories per se are not the morally relevant endpoint of consciousness but insist instead that the global broadcast that normally precedes these things is. But if so, we have to explain why global broadcasting is somehow fundamentally different from more local broadcasting that the subsystems do in smaller regions. After all, the `global broadcasting' that happens in most of our brains usually stays there, rather than being distributed all across planet Earth, yet it still matters to us.

Finally, as \citet{schwitzgebel2012if}'s example of a conscious United States illustrates, the boundaries of where an agent begins or ends aren't necessarily sharp. Is the United States a separate organism from Canada, even if they engage in trade and cross-border migration? And what about Europe when people travel for vacation? Similar kinds of delineation issues arise in the context of simple RL agents: Which parts of the code are the `agent,' and which are the `environment'? It's not always clear, especially if the program is written in a single series of imperative statements without object-oriented organisation. Even for the case of people, our minds are hooked up to our bodies, which are heavily influenced by external objects in our surroundings. At what point does `ourself' end and `the external world' begin? There's not a hard separation; we are all fundamentally part of the same big system. When we talk about different entities, what we're actually doing is carving out conceptual boundaries around parts that are relatively connected and stable, in order to help us conceptualise and describe what's going on. We can do this `carving out' process for RL agents while also recognising that they are part of a bigger, unified program, which may also matter in its own right.

\section*{Implication: Present-day artificial RL deserves a tiny bit of moral consideration}

Contemporary artificial RL agents do not implement most of the functionality of human brains, or even, say, insect brains. But they do contain an important component of what drives goal-directed, welfare-relevant cognition in animals, namely RL, and they have traces of other morally salient characteristics, like emotion (in the form of their computing the reward function based on inputs) and consciousness (such as by broadcasting information updates). Programs equipped with RL have enough of these traits to act successfully in their virtual or physical worlds, showing that they are complete, if limited, agents.

If RL computations do matter at least a tiny bit, the next question is how much they matter relative to other priorities. At the moment I think they rank reasonably low on the list. For instance, fruit flies display rather complex brains compared with many current RL agents. Fruit flies demonstrate RL (e.g., \citet{tempel1983reward}). They have ~100,000 neurons, of which ~200 contain dopamine \citep{whitworth2006drosophila}. In addition, \citet{swinderen2005remote} suggests they may have `the remote roots of consciousness.' Fruit flies are sufficiently intelligent to engage in all the necessary behaviours required for reproduction, repeatedly over millions of years.

In view of their greater cognitive functionality and degree of awareness, it's plausible that fruit flies matter, say, thousands of times more than present-day RL algorithms per learning update. (I'm making up this number, but it seems plausible given 100,000 fruit-fly neurons and the fact that an RL agent is more functionally complex than just one or two neurons.) On the other hand, computers can run thousands of learning updates for artificial agents in the time it would take a fruit fly to have one update. So it may be that, say, the RL algorithms running on a graduate student's laptop are roughly comparable in importance to one insect. (Of course, this estimate is subject to substantial revision as we learn more, or depending on your ethical viewpoint.) But in total the world contains about ten billion billion insects\footnote{This figure is quoted in dozens of sources -- e.g., \citet{berenbaum1996bugs} -- though I'm unable to find the original calculation.} and not nearly so many AI graduate students, so the welfare of insects is a vastly greater moral concern at this stage. But in the long run, as computing power grows and RL agents become increasingly sophisticated, RL looks set to become a pressing ethical consideration in its own right.

Unfortunately, even the welfare of insects and other invertebrates is not generally seen as a significant ethical issue, though the topic is receiving increasing attention \citep{lockwood1987moral,mather2001animal,mason2011invertebrate}, and methods of pain relief and euthanasia for invertebrates have been recommended \citep{cooper2011anesthesia}.

The analogy with laboratory-animal welfare is helpful, because RL research can be approached using similar frameworks as animal research \citep{winsby2013suffering}. Central principles for the use of experimental animals are the `Three Rs' \citep{russell1959principles}. Applied to RL, they would suggest that researchers
\begin{enumerate}
\item \underline{Replace} the use of RL with other algorithms that less closely resemble an agent undergoing emotionally valenced experiences
\item \underline{Reduce} the number of RL agents used
\item \underline{Refine} RL algorithms to be more humane, such as by
\begin{itemize}
\item using rewards instead of punishments
\item not hooking up RL algorithms to higher-level cognitive and emotional faculties
\item running fewer biologically inspired RL algorithms (like TD actor-critic value-function learning) and instead more abstract mathematical ones?
\end{itemize}
\end{enumerate}

It's not clear whether or how much to weigh different algorithms based on their biological plausibility. A very parochial view is to say that we only care about minds that are very similar to ours, including in their algorithmic constitution. So, for instance, if humans don't use policy-gradient learning, then policy-gradient artificial RL would not be ethically significant. A more cosmopolitan view is to not focus so much on the specific algorithm, so long as it gives rise to comparable behaviour and adaptability to the world. In the extreme case, the cosmopolitan view might entail giving ethical consideration to giant lookup tables \citep{block1981psychologism}, though in practice such brains are unlikely to be very common.

So it's debatable how much mileage we can get by refining the type of RL algorithm used. Perhaps the more urgent form of refinement than algorithm selection is to replace punishment with rewards within a given algorithm. RL systems vary in whether they use positive, negative, or both types of rewards:
\begin{itemize}
\item In certain RL problems, such as maze-navigation tasks discussed in \citet{sutton1998introduction}, the rewards are only positive (if the agent reaches a goal) or zero (for non-goal states).
\item Sometimes a mix between positive and negative rewards\footnote{As \citet{barto1983neuronlike} note, `negative reinforcement' in behaviourist psychology refers to reinforcing actions that remove an unpleasant stimulus, such as taking drugs to reduce painful withdrawal symptoms \citep{flora2004power}. What I refer to by `negative reward value' in an RL context could be used in learning either what behaviourists call `negative reinforcement' (which increases inclination to take an action that removes an unpleasant stimulus) or `positive punishment' (which decreases inclination to take an action that causes an unpleasant stimulus). A parallel situation applies for `positive reinforcement' and `negative punishment.' I've avoided using the phrases `negative reinforcement' and `positive reinforcement' in this article to reduce confusion, but when I speak of `negative rewards,' all I mean are reward values that are negative numbers ($r < 0$), and positive numbers for `positive rewards' ($r > 0$), without intending to suggest behaviourist connotations.} is used. For instance, \citet{mccallum1993overcoming} put a simulated mouse in a maze, with a reward of 1 for reaching the goal, -1 for hitting a wall, and -0.1 for any other action.
\item In other situations, the rewards are always negative or zero. For instance, in the cart-pole balancing system of \citet{barto1983neuronlike}, the agent receives reward of 0 until the pole falls over, at which point the reward is -1. In \citet{koppejan2011neuroevolutionary}'s neuroevolutionary RL approach to helicopter control, the RL agent is punished either a little bit, with the negative sum of squared deviations of the helicopter's positions from its target positions, or a lot if the helicopter crashes.
\end{itemize}

Just as animal-welfare concerns may motivate incorporation of rewards rather than punishments in training dogs \citep{hiby2004dog} and horses \citep{warren2007use,innes2008negative}, so too RL-agent welfare can motivate more positive forms of training for artificial learners. \citet{pearce2004hedonistic} envisions a future in which agents are driven by `gradients of well-being' (i.e., positive experiences that are more or less intense) rather than by the distinction between pleasure versus pain. However, it's not entirely clear where the moral boundary lies between positive versus negative welfare for simple RL systems. We might think that just the sign of the agent's reward value $r$ would distinguish the cases, but the sign alone may not be enough, as the following section explains.

\section*{What's the boundary between positive and negative welfare?}

Consider an RL agent with a fixed life of $T$ time steps. At each time $t$, the agent receives a non-positive reward $r_t \leq 0$ as a function of the action $a_t$ that it takes, such as in the pole-balancing example. The agent chooses its action sequence $(a_t)_{t=1...T}$ with the goal of maximising the sum of future rewards:
\begin{equation*}
\sum_{t=1}^T r_t(a_t).
\end{equation*}
Now suppose we rewrite the rewards by adding a huge positive constant $c$ to each of them, $r'_t = r_t + c$, big enough that all of the $r'_t$ are positive. The agent now acts so as to optimise
\begin{equation*}
\sum_{t=1}^T r'_t(a_t) = \sum_{t=1}^T \left( r_t(a_t) + c \right) = Tc + \sum_{t=1}^T r_t(a_t).
\end{equation*}
So the optimal action sequence is the same in either case, since additive constants don't matter to the agent's behaviour.\footnote{The same would also be true if the agent optimised discounted future rewards over a fixed finite or infinite lifetime.

Also, there could be rare cases where behaviour is not identical if the environment depends on the numerical reward values. For example, suppose a robot prints out its last numerical reward to an observing roboticist. If the roboticist sees a positive number, he smiles, and the robot's image sensors detect this as the `roboticist is happy' state. If the roboticist sees a negative number, he frowns, and the robot enters the `roboticist is unhappy' state. Dependence of the environment on the literal reward values is not typical, especially for simple systems like the pole-balancing agent.} But if behaviour is identical, the only thing that changed was the sign and numerical magnitude of the reward numbers. Yet it seems absurd that the difference between happiness and suffering would depend on whether the numbers used by the algorithm happened to have negative signs in front. After all, in computer binary, negative numbers have no minus sign but are just another sequence of 0s and 1s, and at the level of computer hardware, they look different still. Moreover, if the agent was previously reacting aversively to harmful stimuli, it would continue to do so. As Lenhart K. Schubert explains:\footnote{This quotation comes from spring 2014 lecture notes (\url{http://www.cs.rochester.edu/users/faculty/schubert/191-291/lecture-notes/23}, accessed March 2014) for a course called `Machines and Consciousness.'}
\begin{quote}
If the shift in origin [to make negative rewards positive] causes no behavioural change, then the robot (analogously, a person) would still behave as if suffering, yelling for help, etc., when injured or otherwise in trouble, so it seems that the pain would not have been banished after all!
\end{quote}

So then what distinguishes pleasure from pain? Why do I feel that pain has a different emotional texture than pleasure, rather than both feelings lying on a single scale of valuation?

One possibility is that the `hedonic zero point' is determined by whether I would prefer to have a given experience rather than nothing. The RL agent that we considered in the above example had a fixed lifetime of $T$, but if it had a variable lifetime, then its actions would depend substantially on whether the $r_t$ values were positive or negative. If they were negative, the agent would seek to end its life (`commit suicide') as soon as possible.\footnote{Actually, this would depend on the initial value given to the death state for the agent. Since death is an absorbing state after which no further learning happens, the agent can't empirically update its value for the death state. If the initial value was 0, the agent would seek death if its life was full of negative rewards.} If they were positive, it would seek to live as long as it could, because this would make the sum of rewards larger.

This explanation may sound plausible due to its analogy to familiar concepts, but it seems to place undue weight on whether an agent's lifetime is fixed or variable. Yet I would still feel pain and pleasure as being distinct even if I knew exactly when I would die, and a simple RL agent has no concept of death to begin with. 

A more plausible account is that the difference relates to `avoiding' versus `seeking.' A negative experience is one that the agent tries to get out of and do less of in the future. For instance, injury should be an inherently negative experience, because if repairing injury was rewarding for an agent, the agent would seek to injure itself so as to do repairs more often. If we tried to reward \textit{avoidance} of injury, the agent would seek dangerous situations so that it could enjoy returning to safety.\footnote{This example comes from Lenhart K. Schubert's spring 2014 lecture notes (\url{http://www.cs.rochester.edu/users/faculty/schubert/191-291/lecture-notes/23}, accessed March 2014) for a course called `Machines and Consciousness.' These thought experiments are not purely academic. We can see an example of maladaptive behaviour resulting from an association of pleasure with injury when people become addicted to the endorphin release of self-harm.} Injury needs to be something the agent wants to get as far away from as possible. So, for example, even if vomiting due to food poisoning is the best response you can take given your current situation, the experience should be negative in order to dissuade you from eating spoiled foods again.

Still, the distinction between avoiding and seeking isn't always clear. We experience pleasure due to seeking and consuming food but also pain that motivates us to avoid hunger. Seeking one thing is often equivalent to avoiding another. Likewise with the pole-balancing agent: Is it seeking a balanced pole, or avoiding a pole that falls over?

In animal brains, we may be able to tease out some of the distinction between seeking and avoiding at a physiological level. \citet{daw2002opponent} review evidence that humans have two \textit{separate} motivational systems, one appetitive and one aversive. While dopamine is associated with approach, serotonin is associated with inhibition (among many other things). In AI, RL uses a single scalar reward-prediction error $\delta$, which may be positive or negative with any magnitude, but in the brain, firing rates can only be positive, so presumably a different signal (possibly serotonin) is needed to encode significantly negative errors \citep{daw2002opponent}. It's true that dopamine has a baseline firing rate, and when expected rewards are omitted, dopamine firing drops below baseline, but the magnitude of this effect doesn't seem sufficient on its own. Based on these lines of reasoning, \citet{daw2002opponent} develop a computational model in which serotonin acts as the opponent to dopamine:
\begin{equation*}
\textrm{total prediction error} = \textrm{dopamine} - \textrm{serotonin}.
\end{equation*} 
The model is consistent with findings that serotonin is associated with harm avoidance \citep{cloninger1986unified,hansenne1999harm}. \citet{daw2002opponent} also explain how serotonin can help implement an RL system designed to optimise long-run \textit{undiscounted} average reward \citep{mahadevan1996average}, and their model accounts for the influential opponent-process theory of motivation in psychology \citep{solomon1974opponent}. Because serotonin in this model is hypothesised to encode a running-average reward as opposed to current reward, and dopamine is hypothesised to encode a running-average punishment as opposed to current punishment, \citet{daw2002opponent}'s account also explains why dopamine is observed to rise in response to aversive events.

A final explanation for why pain feels different from pleasure may be that the emotional texture of experiences varies based on the pattern of other neural processes that go on when the experience is triggered. Even among negative experiences, we can distinguish among physical pain, depression, fear, embarrassment, guilt, and so on. Each has its own distinct character based on the orchestra of other cognitive instruments that are playing when it happens. \citet{aldridge2010neural} suggest a similar idea for the case of positive experiences:
\begin{quote}
Much of human pleasure has cognitive qualities that infuse uniquely human properties, and it is likely that abstract or higher pleasures depend on cortical brain areas for those qualities. [...T]he particular pattern of coactivated cortical circuits would resolve the high level cognitive features of a pleasantness gloss on sensations or actions.
\end{quote}

Where does all of this leave our pole-balancing agent? Does it suffer constantly, or is it enjoying its efforts? Likewise, is an RL agent that aims to accumulate positive rewards having fun, or is it suffering when its reward is suboptimal? Of course, as with sentience itself, our evaluations of the emotional valences of these cases are up to us to decide, but our uncertainty in how to make this choice is a reason to exercise caution before we run vast numbers of RL computations -- perhaps even those that only use positive rewards ($r > 0$).

While hedonic setpoints vary among humans, with some people enduring chronic depression and others enjoying frequent satisfaction, this fact presents somewhat less of a puzzle than we have with the pole-balancing agent, because depressed humans behave differently than happy ones, whereas the pole balancer behaves exactly the same with a uniform shift in its reward values. For more complex, human-like agents, if they behave similarly to depressed people, perhaps this is an indication of net suffering, and the opposite if they behave similarly to happy people. But it's dubious to extend this heuristic much beyond the realm of agents with close resemblance to mammals. Instead, we need to develop more general principles.

\section*{Commercial applications of RL}

While many state-of-the-art RL systems currently dwell in academia, in the long run I expect most RL computations to happen in the industrial and consumer domains, once technologies using RL become commercialised.

RL has been proposed for many purposes, including
\begin{itemize}
\item playing backgammon \citep{tesauro1994td} and Othello \citep{van2008application}
\item elevator scheduling \citep{barto1996improving}
\item job scheduling \citep{aydin2000dynamic}
\item task scheduling in petroleum production \citep{aissani2009dynamic}
\item web spidering \citep{rennie1999using}
\item stock trading \citep{lee2007multiagent}
\item optimising drug delivery \citep{gaweda2006model,malof2011optimizing}
\item military simulations \citep{sidhu2006hierarchical,collins2013applying,papadopoulos2013behavior}.
\end{itemize}

One of the fields most closely tied with RL is robotics, because it deals with autonomous agents that need to act in the world. In fact, one review article suggested: `The relationship between [robotics and RL] has sufficient promise to be likened to that between physics and mathematics' \citep{kober2013reinforcement}.

Video games may be another hotbed of RL in the future, since RL offers the promise of creating more realistic non-player characters (NPCs). Currently many `game AIs' use hard-coded rules, but these require effort to build, and machine-learning techniques like RL offer the prospect of automating and refining NPC behaviour \citep{patel2011tuning}. The topic has attracted much academic interest \citep{bjornsson2008efficient,amato2010high}. One popular example of RL for video games is learning to play Super Mario \citep{karakovskiy2012mario}. RL has also been applied to the widely studied Open Racing Car Simulator \citep{loiacono2010learning}.

RL in video games presents one of the clearest cases of ethical concern, because the games are visually compelling and many times violent, making it relatively easier to evoke our emotional sympathies. RL has been suggested for first-person shooter games, and a natural way to train enemy NPCs is to inflict punishment on them when they're killed. For instance, \citet{mcpartland2011reinforcement} report their design: `A large penalty (-1.0) was given when the bot was killed, and a small penalty (-0.000002) was given when the bot was wounded.' As RL is increasingly applied in video games, and as the AI algorithms involved become increasingly lifelike, the ethical questionableness of punishing NPCs will grow.

We can imagine some mitigation proposals, along the lines of the Three Rs discussed previously, that would allow gamers to enjoy greater NPC intelligence without quite so much ethical concern. For instance, if the NPCs can be trained extensively offline, so that in the video game they only execute their previously learned rules rather than continuing to learn on the fly during game play, this would reduce the amount of RL required. In games such as \textit{Creatures} \citep{grand1997creatures}, where players can choose how much to punish their AI pets, game designers could build in limitations on the amount of suffering the AIs could endure before they faint, or die, or otherwise terminate the negative input processes. Perhaps the video-game industry could develop protocols for humane game design, pushed along by government regulation or voluntary standards. That said, this might be a challenging proposition, considering that many people already think that exposure to video-game violence is wrong, while the gaming industry has done little in response. We might also worry whether regulations would drive the inhumane games underground, as being what the `cool, hard-core' gamers play.

\citet{whitby2008sometimes} cites examples in which humans have physically abused robots with which they interacted. This may be troubling, but from the perspective of machine welfare, our moral evaluation depends (ignoring instrumental considerations) on whether the robots were wired to respond aversively to the damage they underwent. Moreover, I think the vast majority of potential suffering that robots and other RL agents will experience in the future will not be due to abuse by angry human owners but rather will be built into their utility functions and will result from `natural' interactions with the environment. While perhaps less emotionally salient to observers, this systemic suffering will be far more common, and insofar as it will be at least somewhat preventable, it deserves ethical priority.

It may be easiest to engender concern for RL when it's hooked up to robots and video-game characters because these agents have bodies, perhaps including faces that can display their current `emotional states.' In fact, interacting with another agent, and seeing how it behaves, can incline us toward caring about it whether it has a mind or not. For instance, children become attached to their dolls, and we may sympathise with cartoon characters on television. In contrast, it's harder to care about a batch of RL computations with no visualization interface being performed on some computing cluster, even if their algorithms are morally relevant. It's even harder to imagine soliciting donations to an advocacy organisation -- say, People for the Ethical Treatment of Reinforcement Learners -- by pointing to a faceless, voiceless algorithm. Thus, our moral sympathies may sometimes misfire, both with false positives and false negatives. Hopefully legal frameworks, social norms, and philosophical sophistication will help correct for these biases.

Some feel that placing special emphasis on those we're close to is not a bias but a feature of their moral frameworks. For instance, \citet{coeckelbergh2010robot} proposes a social-relational paradigm for robot ethics based around `relations between various entities, human and non-human, which are inter-dependent and adapt to one another.' This brings robots into the ethical picture `provided that they participate in the social life.' Sadly, such an approach gives less weight to vastly greater numbers of RL agents that may suffer invisibly in back-end industrial computation centres.

The applications of RL in gaming, robotics, and industry are manifold, and they seem likely to expand in the coming decades. That said, these uses of RL are relatively minor compared with what we might anticipate in the far future, if humanity or non-human AIs expand into the galaxy, harnessing the energy of stars to create prodigious amounts of computing power, and requiring massive numbers of robots and other, possibly RL-based agents as workers. The ethical risks in scenarios like these are, to borrow a double entendre from \citet{bostrom2003astronomical}, `astronomical.'

\section*{Do non-RL agents matter?}

If a primary evolutionary purpose of pleasure and pain is to serve as the reward/punishment values in an RL system, do organisms lacking RL not experience pleasure or pain? Perhaps this is one reason why, when scientists ask questions like `Do bugs feel pain?', they look for abilities like RL beyond mere reflex behaviour \citep{efsa2005aspects}.

Are there other features of an organism that matter ethically besides RL? What if it's an apparently goal-directed agent exhibiting complex but not adaptive behaviours, such as NPCs in most modern video games that run using fixed if-then rules (analogous to stimulus-response reflexes in animals) and non-RL algorithmic methods like pathfinding algorithms?

Many industrial-control systems, including simple thermostats, likewise respond to environmental conditions by following pre-programmed rules rather than learning the rules. An RL agent could \textit{become} a thermostat, learning to turn on the heat when it entered the `cold' state and turn on the cooling when it entered the `hot' state. Once trained, the RL agent might act just like the thermostat. But the thermostat didn't have the training phase.

It seems plausible to care about goal-directed agents even if they didn't have a training phase. After all, they still appear to have what we would call preferences; things can still be better or worse for them. When an AI NPC in a first-person shooter is killed, it still fails to accomplish what it was striving for, even if that striving was only being executed by pre-programmed rules.

\citet{torrance2000towards} suggests that even if AIs aren't sentient, one might still value them ethically for attributes like autonomy, intelligence, and cognitive sophistication. My discussion here is similar, except that on my view of sentience, those attributes might indeed be rightly considered part of what makes an agent sentient. But fundamentally it doesn't matter whether we call these criteria part of sentience or part of moral valuation beyond sentience, because the intuitions and conclusions seem to be similar.

I think sufficiently complex rule-based agents probably do have ethical importance, even if they don't perform RL. To pump this intuition, imagine if you took an elderly person and disrupted her brain's RL modules. She wouldn't update her action-value estimates, but she would continue to operate with her existing, well trained estimates. Presumably this person would still seem fairly normal, someone whom we could be friends with and who could tell us about her experiences, at least for a while until the inability to update action tendencies started to cause problems. I would still care a lot about such a person. And in fact, this example may not be purely imaginary. Parkinson's disease is marked by loss of dopamine \citep{kish1988uneven}. This impairs performance on prediction and reinforcement-learning tasks \citep{knowlton1996neostriatal,frank2004carrot}.

It may be that our experience architecture is built at least partly for the purpose of RL, but this doesn't mean that if you eliminate RL, you eliminate experiences. The reward/punishment signals can keep coming, even if the dopamine or other neurochemicals that encode prediction errors stop working. But if we can care about hypothetical humans whose RL abilities have been removed, why not also care about video-game NPCs that act in a goal-directed fashion without any RL training? If it helps to arouse our sympathies, we could imagine training them with RL and then turning the RL off.

Or maybe the latent architecture is also quite relevant. The hypothetical person whose RL capacity was shut down presumably would still have neural systems for input valuation -- for turning signals from the environment and other brain regions into broadcasts that `This feels good' or `This feels bad.' A video-game AI using fixed if-then rules does not have an overt valuation function. That said, some planning agents do explicitly optimise a sum of future rewards even though they don't update actions using reinforcement learning (e.g., \citep{liu2010combining}). Indeed, many kinds of optimisation processes can be seen as choosing actions to increase rewards relative to some reward function.

Note that if we extend ethical significance to goal-directed agents even if they don't use RL, our moral circle of concern expands very wide. Query optimisers, path planners, machine-translation systems, and many other routine computer programs make choices with an eye to optimising a goal function. We can even see this kind of process throughout physics, such as when a protein folds so as to minimise energy \citep{wales2003energy} or when particles choose a movement trajectory so as to minimise `action' \citep{gray2009principle}.

Still, even if our ethical valuation assigns nonzero moral weight to these things, the weight can be exceedingly minuscule, so maybe the practical implications are not as drastic as they might seem. In general, deciding how much to value different features of the universe is a challenging enterprise. It requires both the heart, to assess what kinds of entities we feel compassion towards, and the head, to make our intuitions consistent and identify sources of suffering that we might not ordinarily have noticed. This paper has only begun to scratch the surface.

\section*{Robustness to other views on consciousness}

Caring a little bit about RL algorithms seems a natural extension of a graded view of sentience. If subjective experience is a stance we adopt toward physical processes, then processes that have at least minimal degrees of morally relevant characteristics matter a small amount.

But my graded, `phenomenal stance' approach to consciousness is not universally shared. For example, \citet{torrance2000towards} explicitly rejects as absurd the thesis I advanced in Premise 1:
\begin{quote}
How would we be able to tell if an [AI] were genuinely conscious, rather than just behaving outwardly as conscious? One answer links the matter back to ethical judgment: to claim that x is genuinely conscious may be thought to be definitionally dependent upon the adoption of the appropriate moral attitude towards x. But surely my own consciousness is a matter of objective fact, known to me. Your failure to ascribe consciousness to me is not, therefore, a mere matter of your making a certain moral decision; it is factually false.\footnote{
How would I reply to this? I would firstly deny a hard distinction between first- and third-person viewpoints; everything is just a perception of one sort or another, whether of external stimuli or internal brain states. Secondly, even if we do regard first-person experience as a privileged realm of truth, what do we do with it? All we can say is that I'm conscious in this special first-person way. If we refer to consciousness as `this experience I'm having now,' we can say nothing about other minds, whose brain states are not identical with ours. If we define consciousness as `\textit{kind of like} this experience I'm having now in some relevant ways,' then we get into third-person traits of minds, which cluster into fuzzy, non-binary categories.}
\end{quote}

How does my argument fare for those who feel that whether an agent is conscious is binary? Different theories of consciousness will give different answers, but many of them should at least admit the possibility that RL programs \textit{might} be conscious. Perhaps the likelihood is low, but it should be nonzero. In that case, RL programs would still matter ethically at least a tiny bit in \textit{expected value}. The conclusions would then be similar as what I argued for, with `probability of sentience' playing the role that `degree of sentience' had in my discussion.

Of course, if sentience were a factual, binary property rather than a subjective, fuzzy category, then in the long run, once people understood consciousness well enough, they could potentially conclude with high certainty that RL programs didn't suffer (or did, as the case may be). At that point the practical implications might diverge. Until then, it seems that many views warrant at least thinking twice about the ethical implications of large-scale RL, even if it doesn't yet constitute one of the world's most pressing moral problems.

\section*{Acknowledgements}
Carl Shulman first suggested to me the potential ethical relevance of RL and also refined my understanding of consciousness more generally. Thanks also to David Althaus, Nick Bostrom, Mayank Daswani, Oscar Horta, Rupert McCallum, Joe Mela, Jacob Scheiber, Buck Shlegeris, two anonymous reviewers, and several other people for comments on a draft of this piece.

\bibliographystyle{plainnat}
\bibliography{commonbib}   

\end{document}